\documentclass[11pt]{article}
\usepackage{acl2015}
\usepackage{times}
\usepackage{url}
\usepackage{latexsym}
\usepackage{epsfig}
\usepackage{graphicx}
\usepackage{amsmath}
\usepackage{amssymb}
\usepackage{multirow}
\usepackage{afterpage}
\usepackage{expl3}
\ExplSyntaxOn
\newcommand\latinabbrev[1]{
  \peek_meaning:NTF . {
    #1\@}%
  { \peek_catcode:NTF a {
      #1.\@ }%
    {#1.\@}}}
\ExplSyntaxOff

\def\eg{\latinabbrev{e.g}}

\def\ie{\latinabbrev{i.e}}

\title{Generating Multi-Sentence Lingual Descriptions of Indoor Scenes}

\author{Dahua Lin \\
  { Chinese Univ. of Hong Kong} \\
  {\tt\small dhlin@ie.cuhk.edu.hk} \\\And
  Chen Kong \\
  {Carnegie Mellon Univ.} \\
  {\tt\small chenk@cs.cmu.edu} \\\And
    Sanja Fidler\quad\ \  Raquel Urtasun \\
  University of Toronto \\
  {\tt\small  \{fidler,urtasun\}@cs.toronto.edu} \\
  }

\date{}

\begin{document}
\maketitle

\begin{abstract}
This paper proposes a novel framework for generating lingual descriptions of indoor scenes. 
Whereas substantial efforts have been made to tackle this problem, previous approaches focusing primarily on generating a single sentence for each image, which is not sufficient for describing complex scenes.  We attempt to go beyond this, by generating coherent descriptions with multiple sentences. 
Our approach is distinguished from conventional ones in several aspects:
(1) a 3D visual parsing system that jointly infers objects, attributes, and relations; 
(2) a generative grammar learned automatically from training text; and 
(3) a text generation algorithm that takes into account the coherence among sentences.
Experiments on the augmented NYU-v2 dataset show that our framework can generate natural descriptions with substantially higher ROGUE scores compared to those produced by the baseline.
\end{abstract}

\section{Introduction}

Image understanding has been the central goal of computer vision. Whereas a majority of work on image understanding focuses on class-based annotation, we believe, however, that describing an image using natural language is still the best way to show one's understanding. 
The task of automatically generating textual descriptions for images has received increasing attention from both the computer vision and natural language processing communities. 
This is an important problem, as an effective solution to this problem can enable many exciting real-world applications, such as human robot interaction, image/video synopsis, and automatic caption generation. 

\begin{figure}[t]
    \centering
    \vspace{-2mm}
    \includegraphics[width=1\linewidth,bb=0 0 100 200,trim=0 22 0 0,clip=true]{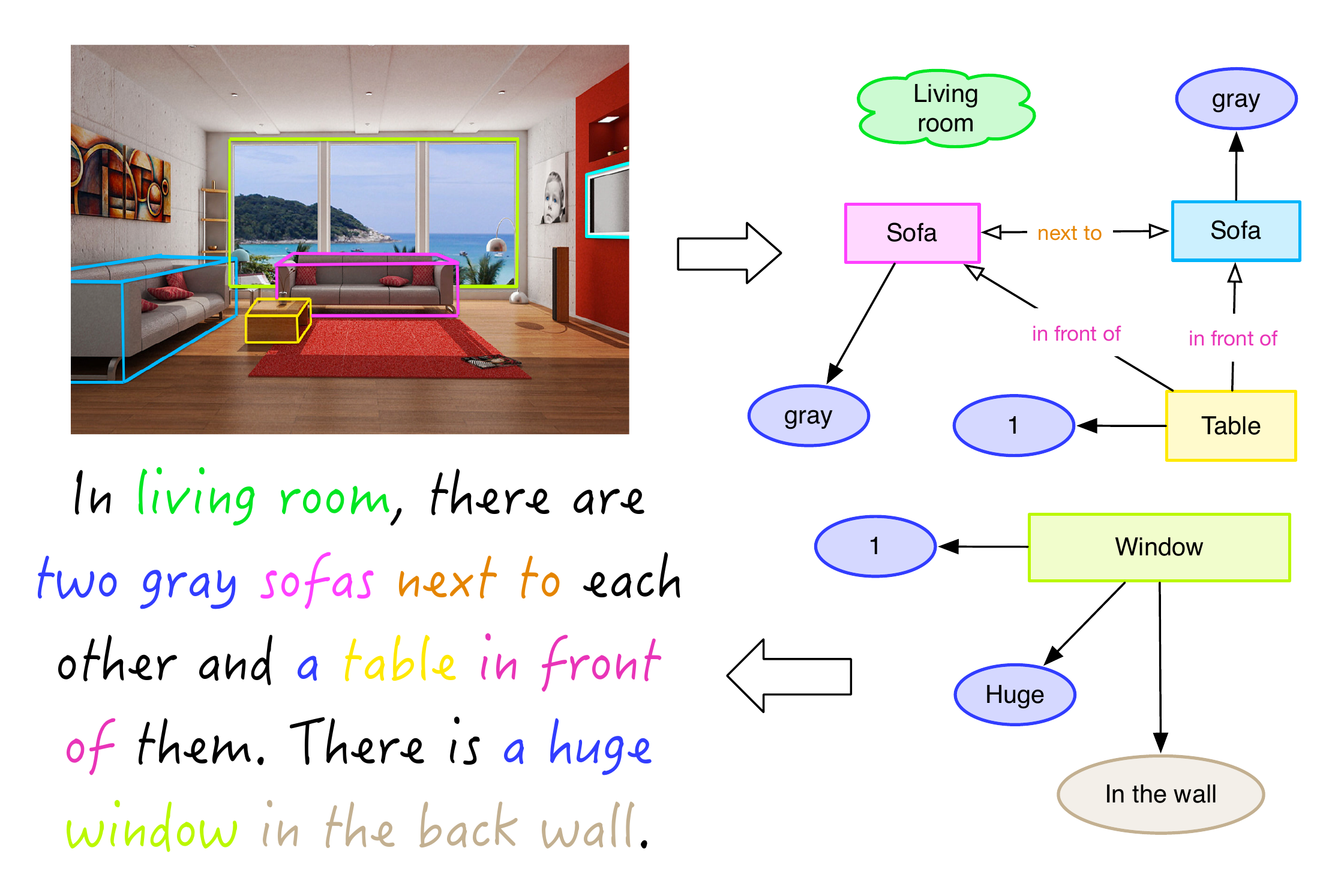}
    \caption{\small Our method visually parses an RGB-D image to get a \emph{scene graph} that represents objects, their attributes and relations between objects, and generates a multi-sentence description via a learned grammar.}
    \vspace{-2mm}
    \label{fig:teaser}
\end{figure}

While this task has been explored in previous work, existing methods mostly rely on pre-defined templates~\cite{SanjaUAI12,krishnamoorthy:aaai13}, which often result in tedious descriptions. Another line of work solves the description generation problem via retrieval, where a description for an image is borrowed from  semantically most similar image from the training set~\cite{Im2Txt,Farhadi10}. This setting is, however, less applicable to complex scenes composed of a large set of objects in diverse configurations, such as for example indoor environments. 

Recently, the field has witnessed a boom in generating image descriptions via deep neural networks~\cite{Kiros14,Karpathy14,Zitnick14} which are able to both, learn a weak language model as well as generalize description to unseen images. These approaches typically represent the image and words/sentences with vectors and reason in a joint embedding space. The results have been impressive, perhaps partly due to powerful representation on the image side~\cite{krizhevsky2012imagenet}. This line of work mainly generates a single sentence for each image, which typically focus on one or two objects and typically contain very few prepositional relations between objects.

In this paper, we are interested in generating multi-sentence descriptions of cluttered indoor scenes, which is particularly relevant for indoor robotics. Complex, multi-sentence output requires us to deal with challenging problems such as consistent co-referrals to visual entities across sentences. Furthermore, the sequence of sentences needs to be as natural as possible, mimicking how humans describe the scene. This is particularly important for example in the context of social robotics to enable realistic communications.   

\begin{figure*}[t]
\vspace{-3.5mm}
    \centering
    \includegraphics[width=0.85\textwidth]{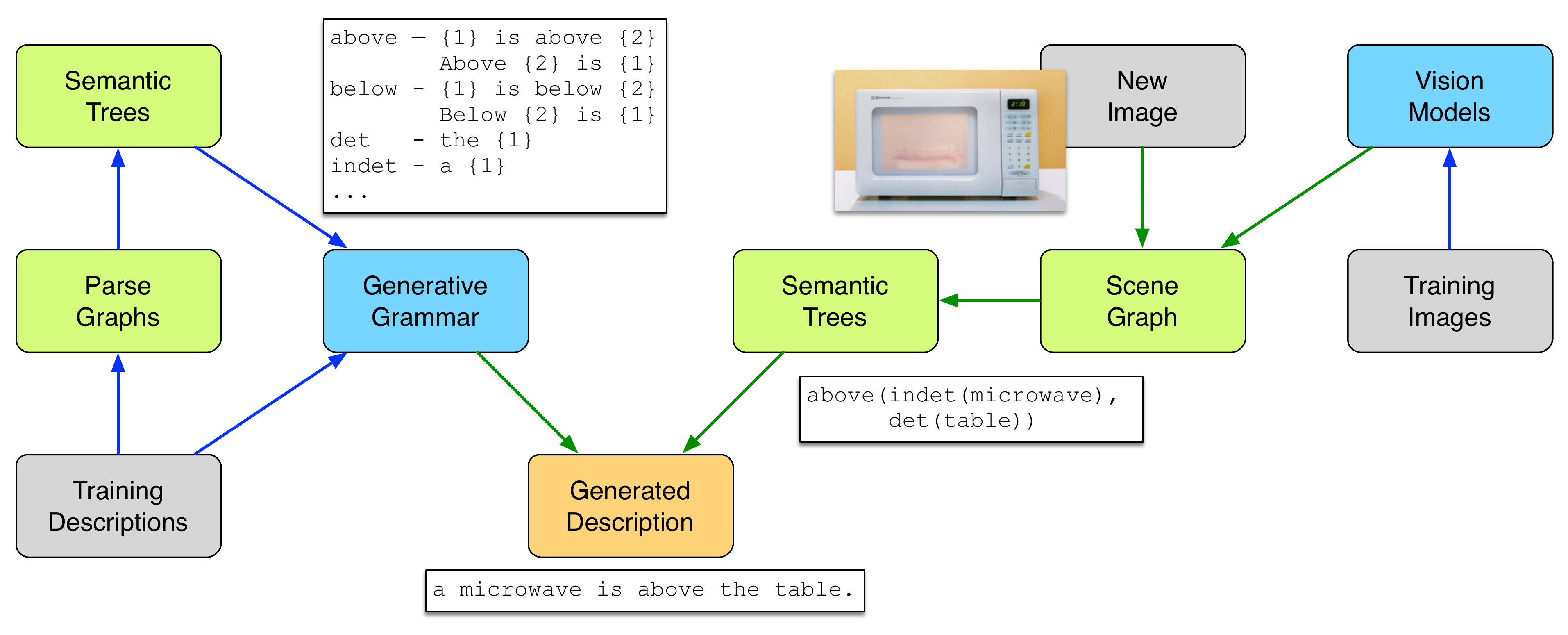}
    \vspace{-3mm}
    \caption{\small The overall framework for description generation. The task consists of the training and the testing phase. In training, the vision models and the generative grammar are respectively learned from a set of RGB-D images and their descriptions. In testing, given a new image, it constructs a scene graph taking into account objects, their attributes and relationships between objects, and transforms it to a series of semantic trees. The learned grammar then generates textual descriptions for these trees.}
    \label{fig:overview}
\end{figure*}

Towards this goal, we develop a framework with three major components: 
(1) a \emph{holistic visual parser} that couples the inference of objects, attributes, and relations to produce a semantic representation of a 3D scene (Fig.~\ref{fig:teaser}); 
(2) a \emph{generative grammar} automatically learned from training text; and 
(3) a \emph{text generation algorithm} that takes into account subtle \emph{dependencies across sentences}, such as logical order, diversity, saliency of objects, and co-references.

To test the effectiveness of our approach, we construct an augmented dataset based on NYU-RGBD~\cite{Silberman12}, where each scene is associated with up to $5$ natural language descriptions from human annotators. This allows us to learn a language model to describe images the way that humans do. 
Experiments show that our method produces natural descriptions, significantly improving the F-measures of ROUGE scores over the baseline. 

\section{Related Work}

A large body of existing work deals with images and text in one form or the other. The dominant subfield exploits text in the form of tags or short sentences as weak labels to learn visual models~\cite{Quattoni07,Li09,Socher10,Gupta08}, as well as attributes~\cite{matuszek12,Silberer13}. This type of approaches have also been explored in videos to learn visual action models from textual summaries of videos~\cite{ramanathan13}, 
or learning visual concepts from videos described with short sentences~\cite{yu13}. 
Another direction is to exploit short sentences associated with images in order to improve visual recognition tasks~\cite{fidler13,KongCVPR14}. 
Just recently, an interested problem domain was introduced in~\cite{malinowski14nips} with the aim to learn how to answer questions about images from Q\&A examples. In~\cite{LinCVPR14}, the authors address visual search with complex natural lingual queries.

There has been substantial work in automatically generating a caption or description for a given image. The most popular approach has been to retrieve a sentence from a large corpus based on similarity of visual content~\cite{Im2Txt,Farhadi10,Kuznetsova12,rohrbach13iccv,Yang11}. This line of work bypasses having to deal with language template specification or template learning. However, typically such approaches adopt a limited image representation such as triplets action-object-scene~\cite{Farhadi10}. This makes a restrictive setting, as neither the image representation nor the retrieved sentence can faithfully model a truly complex scene. In~\cite{Kuznetsova14} the authors go further by only learning phrases from related images.

Parallel to our work, there has been a recent boom in image description generation with deep networks~\cite{Kiros14,Karpathy14,Vinyals14,Mao14,Donahue14,Fang14,Zitnick14}. These methods transform the image as well as a sentence into a vector representation and learn a joint embedding between the two modalities. The output of these approaches is typically a short sentence
for each image. In contrast, our goal here is to generate \emph{multiple} dependent sentences that describe the salient objects in the scene, their properties and spatial relations. 


Generating descriptions has also been explored in the video domain. 
\cite{SanjaUAI12,krishnamoorthy:aaai13} output a video description in the form of subject-action-object. In~\cite{Corso13},  ``concept detectors'' are formed, which are detectors for combined object and action or scene in a particular chunk of a video. Via lingual templates the concept detectors of particular types then produce cohesive video descriptions. Due to a limited set of concepts and templates the final descriptions do not seem very natural. \cite{rohrbach13iccv} predicts semantic representations 
from low-level video features and uses machine translation techniques to generate a sentence. 

The closest to our work is~\cite{BabyTalk,mitchell12,Kuznetsova14} which, like us, is able to describe objects, their modifiers, and prepositions between objects. However, our paper differs from~\cite{BabyTalk,mitchell12} in several important ways. In our work, we reason in 3D as opposed to 2D giving us more natural \emph{physical} interpretations. We aim to describe rich indoor scenes that contain many objects of various classes and appear in various arrangements. 
In such a setting, describing every detectable object and all relations between them as in~\cite{BabyTalk} would generate prohibitively long, complex and unnatural descriptions.
Our model tries to mimic \emph{what} and \emph{how} people describe such complex 3D scenes, thus taking into account visual saliency at the level of objects, attributes and relations, as well as the ordering and coherence of sentences. Another important aspect that sets us apart from most past work is that instead of using a few hand-crafted templates, we \emph{learn} the grammar from training text. 





\section{Framework Overview}

Our framework for generating descriptions for indoor scenes is based on a key rationale: images and their corresponding descriptions are two different ways to express the underlying \emph{common semantics} shared by both. 
As shown in Figure~\ref{fig:overview}, given an image, it first recovers the underlying \emph{semantics} through holistic visual analysis~\cite{Lin13}, which results in a \emph{scene graph} that captures detected objects and the spatial relations between them (\eg~\textit{on-top-of} and \textit{near}, etc).

The \emph{semantics} embodied by a visual scene usually has multiple aspects. When describing such a \emph{complex} scene, humans often use a paragraph comprised of multiple sentences, each focusing on a specific aspect. To imitate this behavior, this framework transforms the \emph{scene graph} into a sequence of \emph{semantic trees}, and yields multiple sentences, each from a \emph{semantic tree}. 
To make the results as natural as possible, we adopt two strategies:
(1) Instead of prescribing templates in advance, we learn the \emph{grammar} from a training set -- a set of RGB-D scenes with descriptions provided by humans. 
(2) We take into account dependencies among sentences, including \emph{logical order}, \emph{saliency}, \emph{coreference} and \emph{diversity}.

\section{From RGB-D Images to Semantics}

Given an RGB-D image, we extract \emph{semantics} through \emph{holistic visual parsing}. Particularly, we first parse the image to obtain the objects of interest, their attributes, and their physical relations, and then construct a \emph{scene graph}, which provides a coherent summary of these aspects.

\subsection{Holistic Visual Parsing}
\label{sub:visual}



To parse the visual scene we use a recently proposed approach for 3D object detection in RGB-D data~\cite{Lin13}. We briefly summarize this approach here. 
First, a set of \emph{``objectness'' regions} are generated following~\cite{Smin12}, which are encouraged to respect intensity as well as occlusion boundaries in 3D. These regions are projected to 3D via depth and then cuboids are fit tightly around them, under the constraint that they are parallel to the ground floor.

A \emph{holistic CRF model} is then constructed to jointly reason about the classes of the \emph{cuboids} as well as the class of the scene (\eg, kitchen, bathroom). The CRF thus has a random variable for each cuboid representing its class, and a variable for the scene. To have the possibility to remove a bad, non-object cuboid, we have an additional background state for each cuboid. The model exploits various geometric and semantic relations by incorporating them into the CRF formulation as \emph{potentials}, which are summarized below:

\textbf{Scene Appearance.} To incorporate global information, a unary potential over the scene label  is computed by means of a logistic on top of the scene classification score~\cite{xiao10}. 

\textbf{Cuboid class potential.} Appearance-based classifiers, including CPMC-o2~\cite{Smin12cls}, superpixel scores~\cite{ren12} are used to classify cuboids into a pre-defined set of object classes. In this paper, we additionally use CNN~\cite{krizhevsky2012imagenet} features for classification. The classification scores for each cuboid are used as different unary potentials in the CRF.

\textbf{Object geometry.} Cuboids are also classified based on geometric features (\eg~\emph{height}, \emph{longer width}, \emph{aspect ratio}, etc) with SVM, and the classification scores used as another unary potential.

\textbf{Semantic context.} Two co-occurrence relationships are used: \emph{scene-object} and \emph{object-object}. The potential values are estimated from the training set by counting the co-occurence frequencies. 

\textbf{Geometric context.} Two potentials are used to exploit the spatial relations between cuboids in 3D, encoding 
\emph{close-to}  and \emph{on-top-of} relations.  The potentials are defined to be the empirical co-occurrence frequencies for each type of relation. 

\vspace{3pt}
The CRF weights to combine the potentials are learned with a primal dual learning framework~\cite{Hazan10}, and inference of class labels is done with an approximated algorithm~\cite{Schwing11}.

\subsection{Scene Graphs}

Based on the extracted visual information, we construct a \emph{scene graph} that captures \emph{objects}, their \emph{attributes}, such as color and size, and the relations between them.
In particular, a \emph{scene graph} uses \emph{nodes} to represent objects and their attributes, and \emph{edges} to represent relations between nodes. 
Here, we consider three kinds of edges: \emph{attribute edges} that link objects to their attributes, \emph{position edges} that represent the positions of objects relative to the scene, (\eg~\emph{corner-of-room}), and \emph{pairwise edges} that characterize the relative positions between objects (\eg~\textit{on-top-of} and \textit{next-to}).

Given an image, a set of objects (with class labels) and the scene class are obtained through visual parsing as explained in the previous Section. However, to form a \emph{scene graph}, we still need further analysis to extract \emph{attributes} and \emph{relations}. For each object we also compute \emph{saliency}, i.e. how likely an object  will be described. We next describe how we obtain such information.


\vspace{-2mm}
\paragraph{Object attributes:} For each object, we use RGB histograms and C-SIFT, and cluster them to obtain a visual word representation. We train classifiers for nine colors that are most mentioned in the training set, 
as well as two material properties (\emph{wooden} and \emph{bright}). We also train classifiers for four different sizes (\emph{wide}, \emph{tall}, \emph{large}, and \emph{small}) using geometric features. To encode the correlations between size and the object class, we augment the feature with a class indicator vector.
%

\vspace{-2mm}
\paragraph{Object saliency:} The dataset of~\cite{KongCVPR14} contains alignment between the nouns in a sentence and the visual objects in the scene. We make use of this information to train a classifier predicting whether an object in the scene is likely to be mentioned in text. We train an SVM classifier using class-based features (classification scores for each cuboid), geometric relations (volume, distance to camera), and color features.

\vspace{-2mm}
\paragraph{Object relations:} We consider six types of \emph{object positions} (\emph{corner-of-room},  \emph{front-of-camera}, \emph{far-away-from-camera}, \emph{center-of-room}, \emph{left-of-room}, and \emph{right-of-room}), and eight types of \emph{pairwise relations} (\emph{next-to}, \emph{near}, \emph{top-of}, \emph{above}, \emph{in-front-of}, \emph{behind}, \emph{to-left-of}, and \emph{to-right-of}). 
We manually specify a few rules that help us decide whether a specific relation is present or not\footnote{We tried obtaining ground-truth for relations via MTurk (which would allow us to train classifiers instead), however, the results of all batches were extremely noisy.}.


\section{Generating Lingual Descriptions}

Given a \emph{scene graph}, our framework generates a descriptive paragraph in two steps.
First, it transforms the scene graph into a sequence of \emph{semantic trees}, each focusing on a certain \emph{semantic aspect}. Then, it produces sentences, one from each semantic tree, following a \emph{generative grammar}.

\subsection{Semantic Trees}

A \emph{semantic tree} captures information such as \emph{what entities are being described} and \emph{what are the relationships between them}. Specifically, a semantic tree contains a set of \emph{terminal nodes} corresponding to individual entities or their attributes and \emph{relational nodes} that express relations among them. Consider a sentence \textit{``A red box is on top of a table"}. The corresponding semantic tree can be expressed as 
\begin{verbatim}
 on-top-of(indet(color(box, red)), 
           indet(table))
\end{verbatim}
This tree has three terminals: \textit{``box"}, \textit{``table"}, and \textit{``red"}. The relation node \textit{``color(box, red)"} describes the relation between \textit{``box"} and \textit{``red"}, namely, \textit{``red"} specifying the color of the \textit{``box"}. The relation \textit{``indet"} qualifies the cardinality of its child; while \textit{``on-top-of"} characterizes the spatial relation between its children.

\subsection{Dependencies among Sentences}

In human descriptions, sentences are put together in a way that makes the resultant paragraphs coherent. In particular, the \emph{dependencies} among sentences, as outlined below, play a crucial role in preserving the coherence a descriptive paragraph:

\textbf{Logical order.} When describing a scene, people present things in certain orders. The leading sentence often mentions the type of the entire scene and one of the most salient object, \eg~\textit{``There is a table in the dining room."}
    
\textbf{Diversity.} People generally avoid using the same prepositional relation in multiple sentences. Also, when an object is mentioned in multiple sentences, it usually plays a different role, \eg~\textit{``There is a table near the wall. On top of the table is a microwave oven.''} 
Here, \textit{``table''} respectively serves as a \emph{source} and a \emph{target} in these two sentences\footnote{Each relation is considered as an edge. For example, in phrases \textit{``A on-top-of B"} and \textit{``A near B"}, \textit{``A"} is considered as the source, while \textit{``B"} considered as the target.}.  
    
\textbf{Saliency.} \emph{Saliency} influences the order of sentences. The statistics in~\cite{KongCVPR14}~shows that bigger objects are often mentioned earlier on in a description and co-referred across sentences, \eg~one would say \textit{``This room has a dining table with a mug on top. Next to the table is a chair.''} and not \textit{``There is a mug on a table. Next to the mug is a chair.''} \emph{Saliency} also depends on context, \eg~for bathrooms, toilets are often mentioned. 
    
\textbf{Co-reference.} When an object is mentioned for the second time following its debut, a pronoun is often used to make the sentence concise. 
    
\textbf{Richness vs. Conciseness.} When talking about an object for the first time, describing its color/size makes the sentence  interesting and informative. However, this is generally unnecessary the next time the object is mentioned.

\subsection{From Scene Graphs to Semantic Trees}

Motivated by these considerations, we devise a method below that transforms a \emph{scene graph} into a sequence of \emph{semantic trees}, each for a sentence.

First of all, we initialize $w^s_i = w^t_i = \mathfrak{s}_i \cdot \mathfrak{c}_i$. 
Here, $w^s_i$ and $w^t_i$ are the weights that respectively control how likely the $i$-th object will be chosen as a \emph{source} or a \emph{target} in the next sentence; $\mathfrak{s}_i$ is a positive value measuring the \emph{saliency} of the $i$-th object, while $\mathfrak{c}_i$ is given by the classifier to indicate its confidence as to whether it makes a correct prediction of the object's class. These weights are updated as the generation proceeds. 

To generate the leading sentence, we first draw a \emph{source} $i$ with a probability proportional to $w^s_i$, and create a semantic tree by choosing a relation, say \textit{``in''}, which would lead to a sentence like \textit{``There is a table in the dining room."} Once the $i$-th object is chosen to be a source, $w^s_i$ will be set to $0$, precluding it from being chosen as a source again. However, $w^t_i$ remains unchanged, as it remains fine for it to serve as a target later.

For each subsequent sentence, we draw a source $i$, a target $j$, and a relation $r$ between $i$ and $j$, with probability proportional to $w^s_i w^t_j \rho_r$, where $\rho_r$ is the prior weight of the relation $r$. 
At each iteration, one may also choose to terminate without generating a new sentence, with a probability proportional to a positive value $\tau$. These choices together result in a semantic tree in the form of \textit{``r(make\_tree(i), make\_tree(j))"}. Here, \textit{``make\_tree(i)"} creates a sub-tree describing the object $i$, which may be \textit{``indet(color(table, black))"} when the color is known.

After the generation of this semantic tree, the weights $w^s_i$, $w^t_j$, and $\rho_r$ will be set to zero to prevent the objects $i$ and $j$ from being used again for the same role, and the relation $r$ from being chosen next time. Our algorithm also takes care of \emph{co-references} --  if an object is selected again in the next sentence, it will be replaced by a pronoun.

\subsection{Grammar and Derivation}

Given a \emph{semantic tree}, our framework produces a sentence following a \emph{generative grammar}, namely, a map from each semantic relation to a set of templates (\ie~derivation rules), as illustrated below:

{\small
\begin{verbatim}
indet     --> a {1} 
color     --> {2} {1} 
on-top-of --> {1} is on top of {2}
              On top of {2} is {1}
              There is {1} on top of {2}
\end{verbatim}
}

Each template has a weight that is set to its frequency in the training set.
The generation of a sentence from a semantic tree proceeds from the root, and downward recursively to the terminals. For each relation node, a template will be chosen, with a probability proportional to the associated weight. Below is an example showing how a sentence is derived following the grammar above.

{\small
\begin{verbatim}
   {on-top-of(indet(color(box, red)), 
              indet(table))}
=> {indet(color(box, red))} is on top of 
   {indet(table)}
=> a {color(box, red)} is on top of a table
=> a red box is on top of a table
\end{verbatim}
}

As the choices of templates for relational nodes are randomized, different sentences can be derived for the same tree, with different probabilities.

\subsection{Learning the Grammar}
\label{sub:learng}

\begin{figure}[t]
    \centering
    \includegraphics[width=0.5\textwidth]{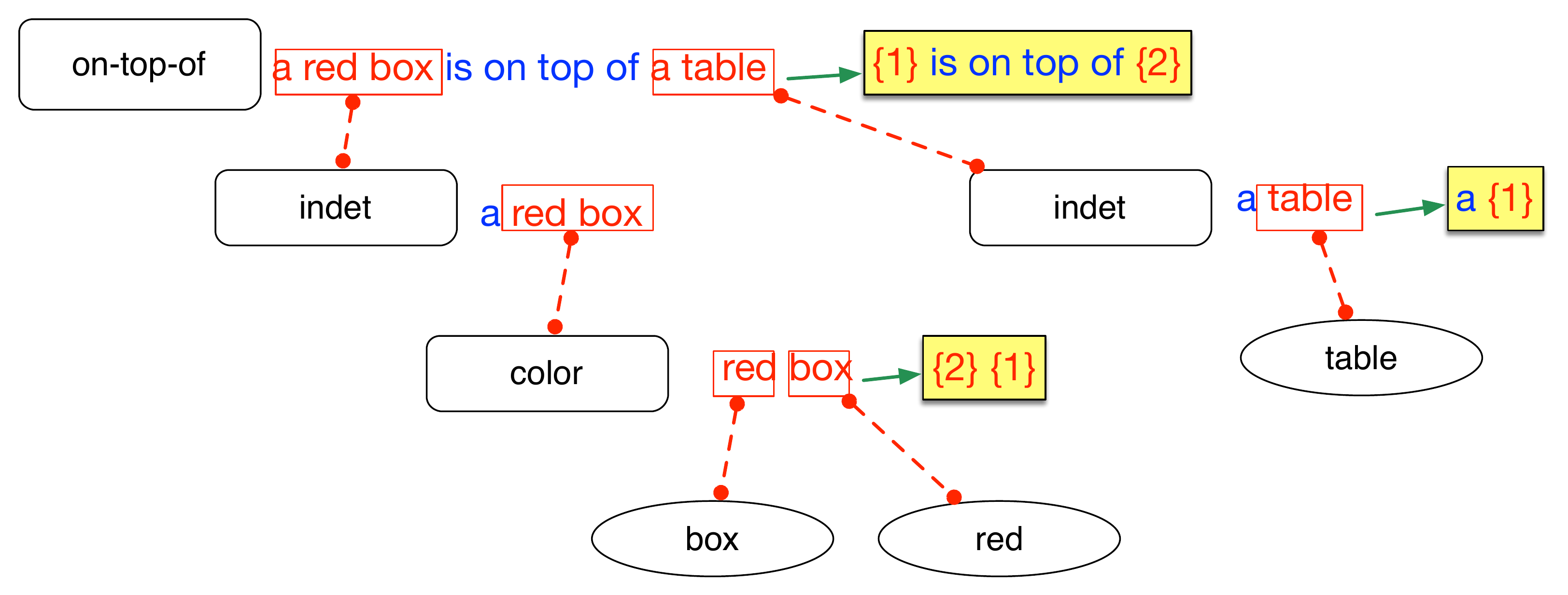}
    \vspace{-8mm}
    \caption{\small The process to derive templates by matching semantic nodes to parts of the sentence. Starting from the root node, the learning algorithm identifies the ranges of words corresponding to the child nodes, and replaces them with a placeholder to obtain a template. This proceeds downward recursively until all relation nodes are processed.}
    \label{fig:learngrammar}
\end{figure}

The \emph{grammar} for generating sentences are often specified manually in previous work~\cite{SanjaUAI12,Corso13}. This way, however, is time consuming, unreliable, and tends to oversimplify the language. In this work, we explore a new approach, that is, to learn the grammar from data. 
The basic idea is to construct a semantic tree from each sentence through linguistic parsing, and then derive the templates by matching nodes of the semantic tree to parts of the sentence. 

First, we use the Stanford parser~\cite{POS} to obtain a \emph{parse tree} for each sentence, which is then simplified through a series of filtering operations. For example, we merge noun phrases (\eg~\textit{``fire distinguisher"}) into a single node and compress common prepositional phrases (\eg~\textit{``in the left of"}) into a single link.

A semantic tree can then be derived by recursively translating the simplified trees. This is straightforward. For example, a noun \textit{``box"} with an adjective \textit{``red"} will be translated into \textit{``color(box, red)"}; a noun with a definite or indefinite article will be translated into an \textit{det} and \textit{indet} relation node; two nouns or noun phrases \textit{``A"} and \textit{``B"} linked by a prepositional link \textit{``above"} will be translated into \textit{``above(A, B)"}. 


With a sentence and a semantic tree constructed thereon, we can derive the template through \emph{recursive matching}, where matched children are replaced by a placeholder, while other words are preserved literally in the template. Figure~\ref{fig:learngrammar} illustrates this procedure. We collect templates respectively for each relation, and set the weight of each template to its frequency. Empirically, we observed a long tailed distribution -- a small number of common templates occur many times, while a dominant portion of templates are used sporadically. To improve the reliability, we discard all the templates that occur less than $5$ times and all relations whose total weight is less than $20$.

\begin{table*}[t]
\vspace{-3mm}
\centering
{\footnotesize
\begin{tabular}{|c|c||ccc|ccc|ccc|}
\hline
\multirow{2}{1.0cm}{objects} & \multirow{2}{1.0cm}{config} & \multicolumn{3}{|c|}{ROUGE1}  &\multicolumn{3}{c|}{ROUGE2} &\multicolumn{3}{c|}{ROUGES} \\
\cline{3-11} &  & {R} & {P} & {F} & {R} & {P} & {F} & {R} & {P} & {F} \\
\hline
\multicolumn{2}{|c||}{baseline} & 0.3000 & 0.2947 & 0.2968& 0.0667 & 0.0657 & 0.0661& 0.1026 & 0.1006 & 0.1014\\
\hline
GT & L0 & 0.3332 & 0.3249 & 0.3281& 0.0786 & 0.0765 & 0.0773& 0.1372 & 0.1334 & 0.1348\\
GT & L1 & 0.3378 & 0.3294 & 0.3327& 0.0838 & 0.0816 & 0.0824& 0.1397 & 0.1359 & 0.1373\\
GT & L2 & 0.3392 & 0.3308 & 0.3340& 0.0849 & 0.0827 & 0.0835& 0.1409 & 0.1370 & 0.1385\\
GT & L3 & 0.3770 & 0.3676 & 0.3712& \bf0.1092 & \bf0.1067 & \bf0.1076& \bf0.1629 & \bf0.1584 & \bf0.1601\\
GT & L4 & \bf0.3775 & \bf0.3680 & \bf0.3716& 0.1064 & 0.1040 & 0.1049& 0.1598 & 0.1554 & 0.1570\\
GT & L5 & 0.3755 & 0.3658 & 0.3695& 0.1008 & 0.0984 & 0.0993& 0.1563 & 0.1519 & 0.1536\\
\hline
Real & L0 & 0.3243 & 0.3161 & 0.3192& 0.0752 & 0.0735 & 0.0742& 0.1306 & 0.1270 & 0.1283\\
Real & L1 & 0.3347 & 0.3266 & 0.3296& 0.0814 & 0.0795 & 0.0802& 0.1362 & 0.1325 & 0.1338\\
Real & L2 & 0.3338 & 0.3256 & 0.3286& 0.0816 & 0.0796 & 0.0803& 0.1356 & 0.1319 & 0.1332\\
Real & L3 & 0.3641 & 0.3541 & 0.3580& 0.1045 & 0.1019 & 0.1029& 0.1546 & 0.1499 & 0.1517\\
Real & L4 & 0.3663 & 0.3560 & 0.3600& 0.1039 & 0.1011 & 0.1022& 0.1534 & 0.1486 & 0.1504\\
Real & L5 & 0.3675 & 0.3570 & 0.3611& 0.1021 & 0.0994 & 0.1004& 0.1526 & 0.1478 & 0.1496\\
\hline
\end{tabular}
\vspace{-5pt}
\caption{\small ROGUE scores for the baseline and our approach under configurations at different levels. Here, ``GT'' and ``Real'' respectively refer to the results obtained based on annotated objects and objects detected by the visual parsing method. For each ROGUE metric, we report the recall (R), precision (P), and F-scores (F) averaged over all scenes and $10$ randomized runs.}
\label{tab2}
}
\end{table*}

\section{Experimental Evaluation}

We test the proposed framework on the NYU-v2 dataset~\cite{Silberman12} augmented with an additional set of textual descriptions, one for each image. Particularly, we focus on assessing both the relevance and quality of the generated descriptions.  

\subsection{Data Preparation}

The NYU-v2 dataset has $1449$ RGB-D images of indoor scenes (\eg~dining rooms, kitchens, offices). These images are divided into a training a testing set, following the partition used in~\cite{Lin13}. The training set contains $795$ scenes, while the testing set contains the remaining $654$. We use the descriptions from~\cite{KongCVPR14} which were collected by asking MTurkers to describe the image to someone who does not see it in order to provide him/her with a vidid impression of the scene. The number of sentences per description ranges from 1 to 10 with an average of 3. There are on average 40 words in a description.

We learn the generative grammar using the algorithm described in Section~\ref{sub:learng} from the training set of descriptions. We also train the CRF for visual analysis and apply it to detect objects and predict their attributes and relations, following the procedure described in Section~\ref{sub:visual}. These models are then used to produce textual descriptions for each test scene.  

\def\IH{5.24cm}
\def\IHW{5.22cm}
\def\IHH{5.2cm}
\begin{figure*}[!htb]
    \centering
    \includegraphics[height=\IH,trim=105 265 105 60,clip=true]{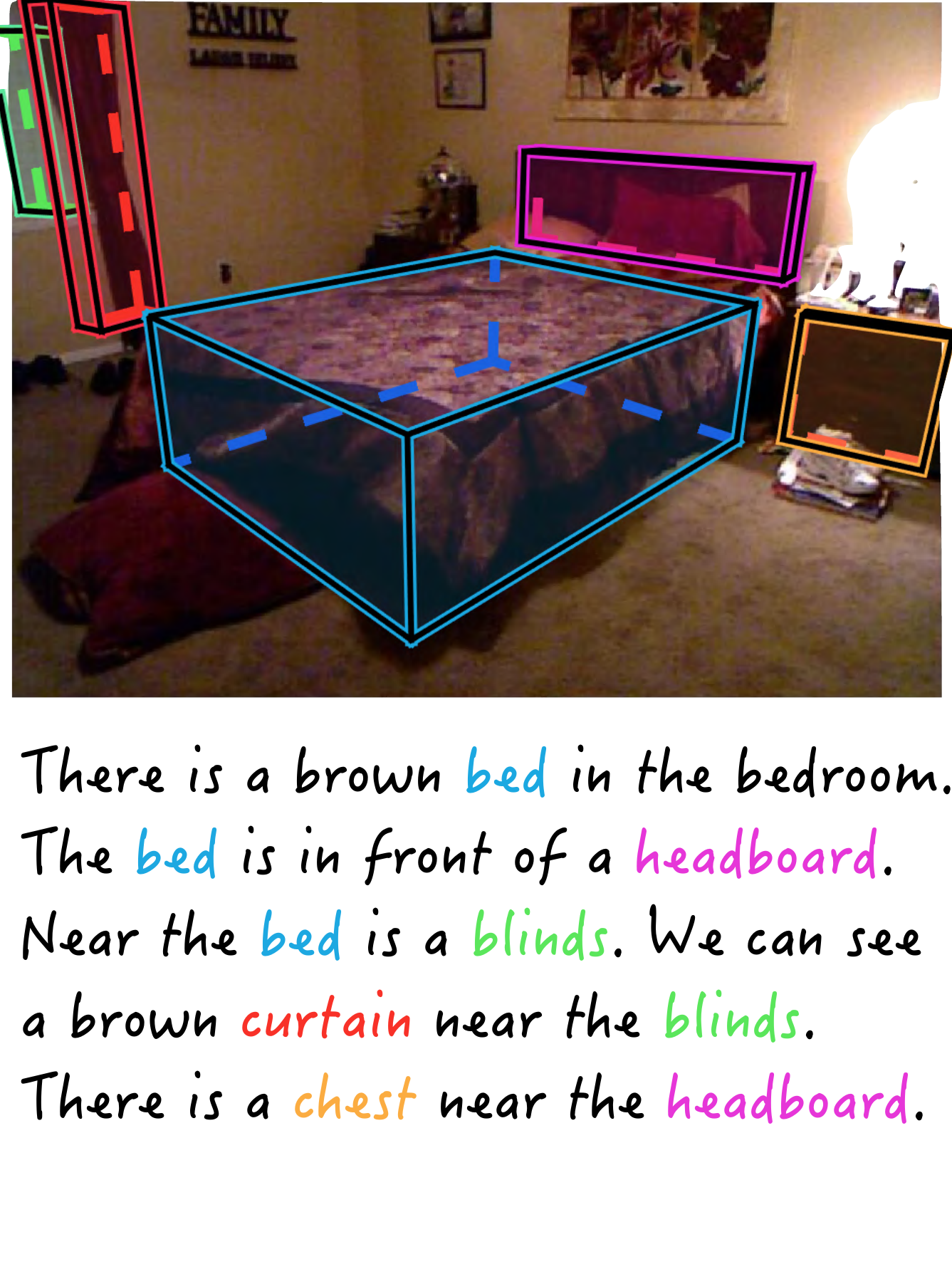}
    \includegraphics[height=\IH,trim=105 265 105 60,clip=true]{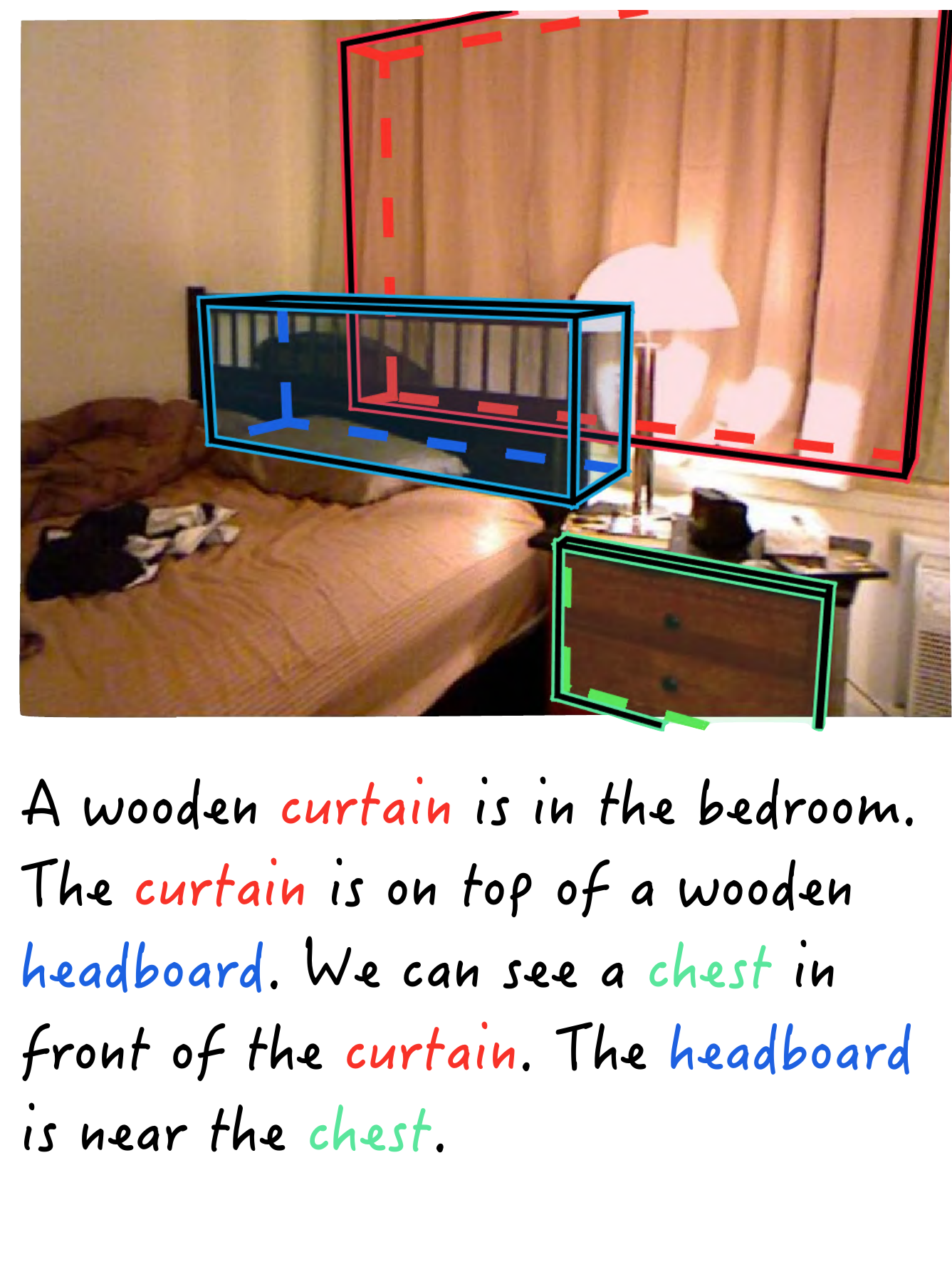}
    \includegraphics[height=\IH,trim=105 265 105 60,clip=true]{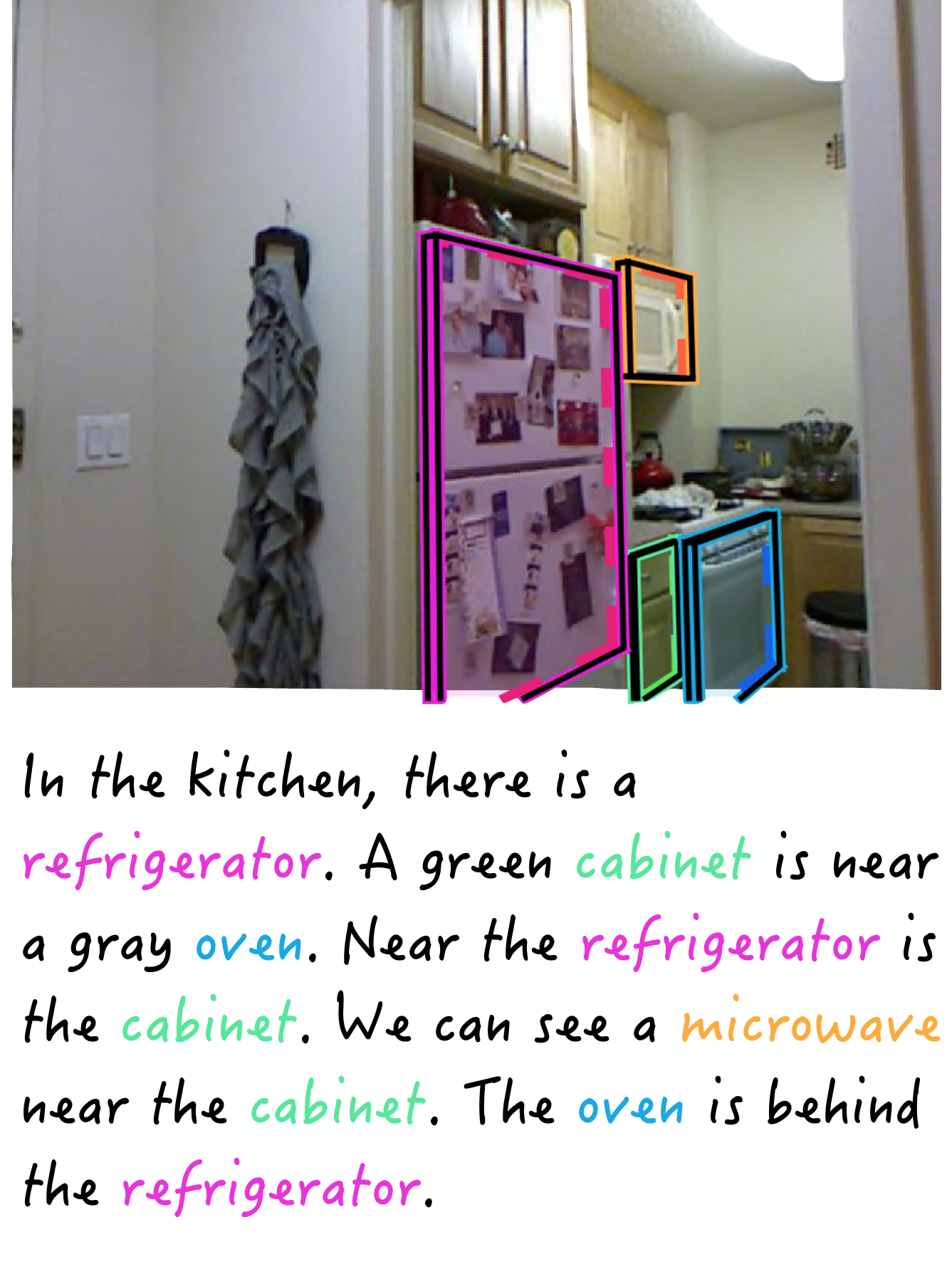}
    \includegraphics[height=\IH,trim=105 265 105 60,clip=true]{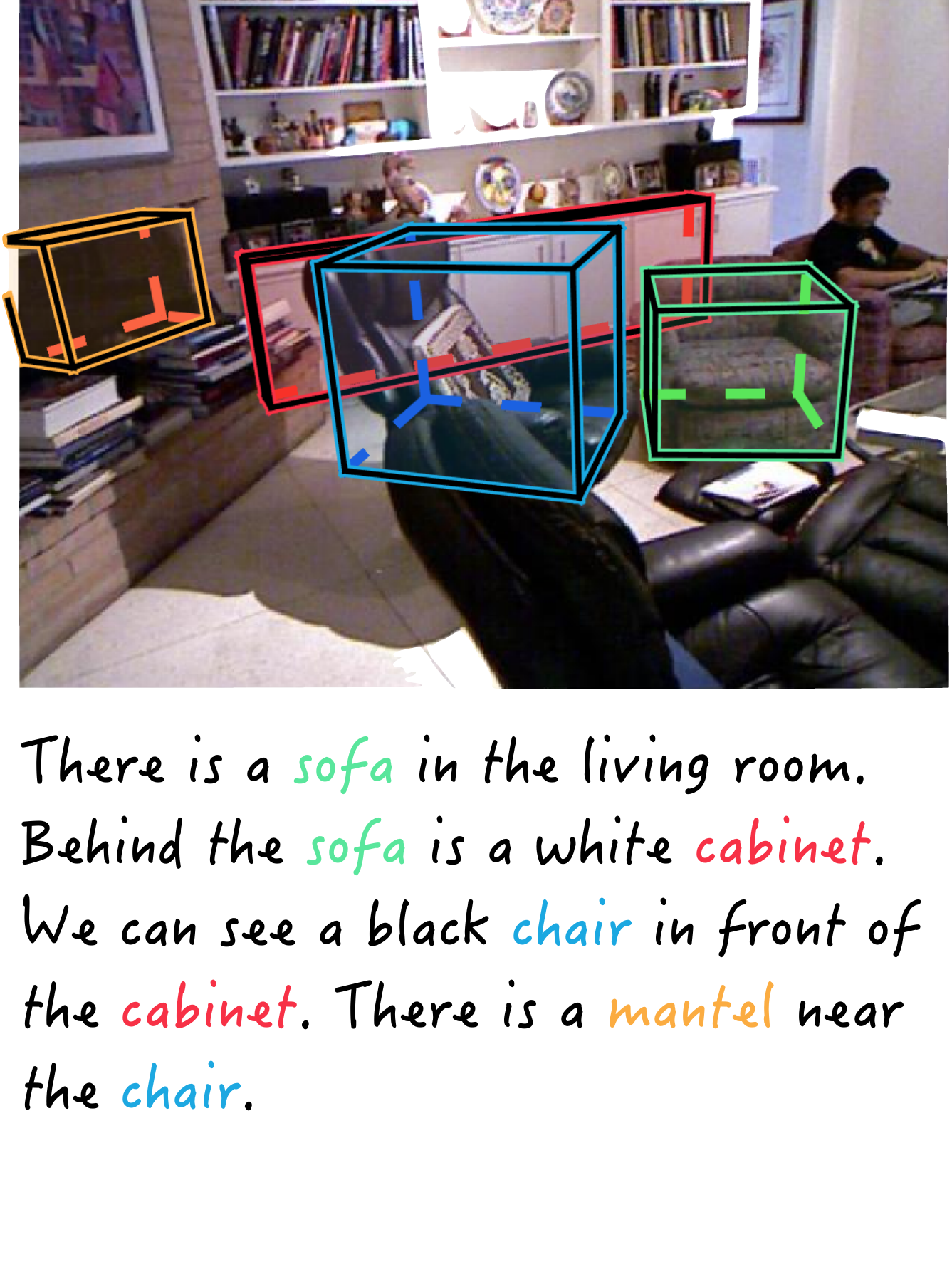}\\[0.1mm]
        \includegraphics[height=\IHW,trim=105 270 105 60,clip=true]{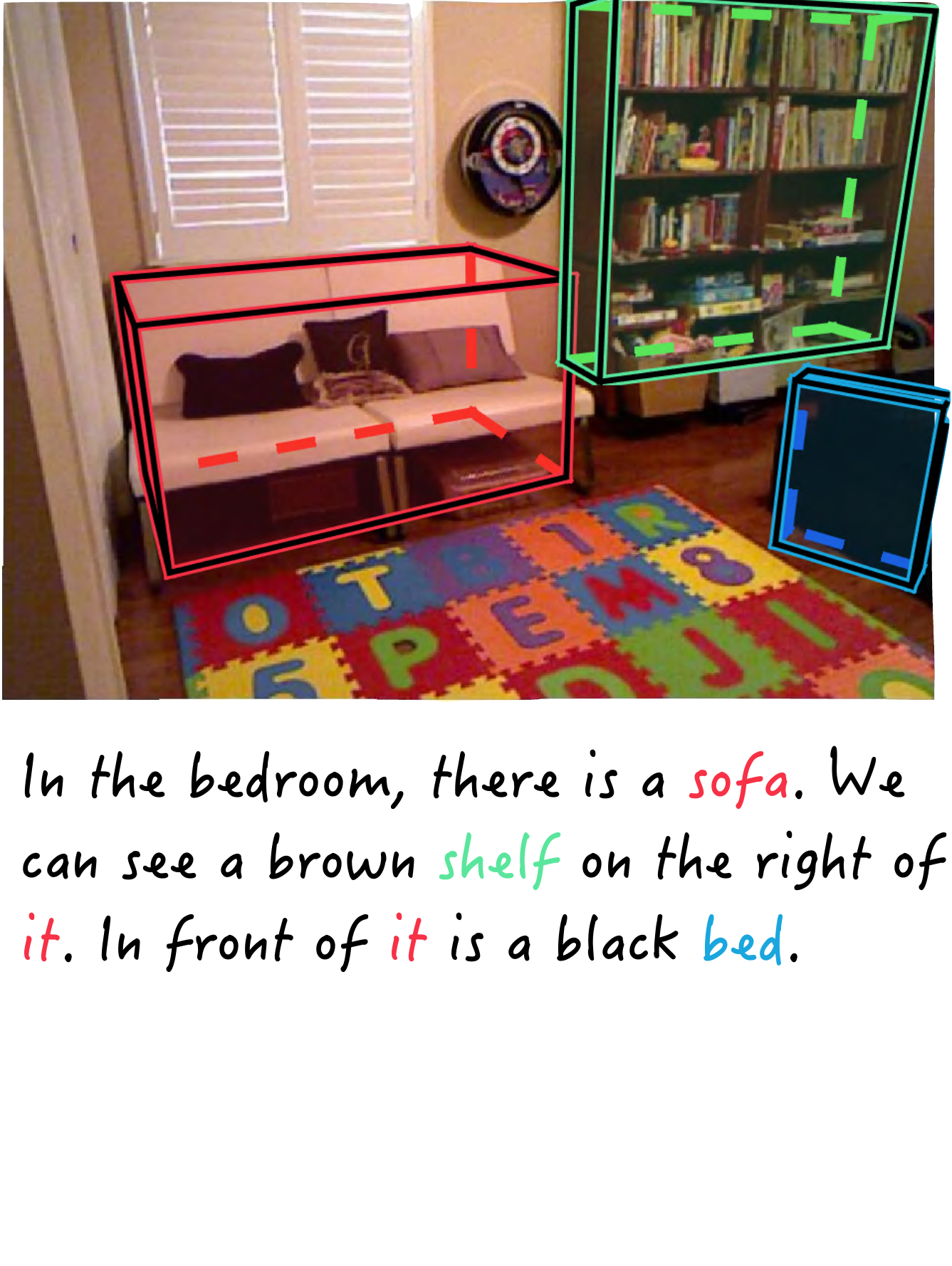}
    \includegraphics[height=\IHW,trim=105 270 105 60,clip=true]{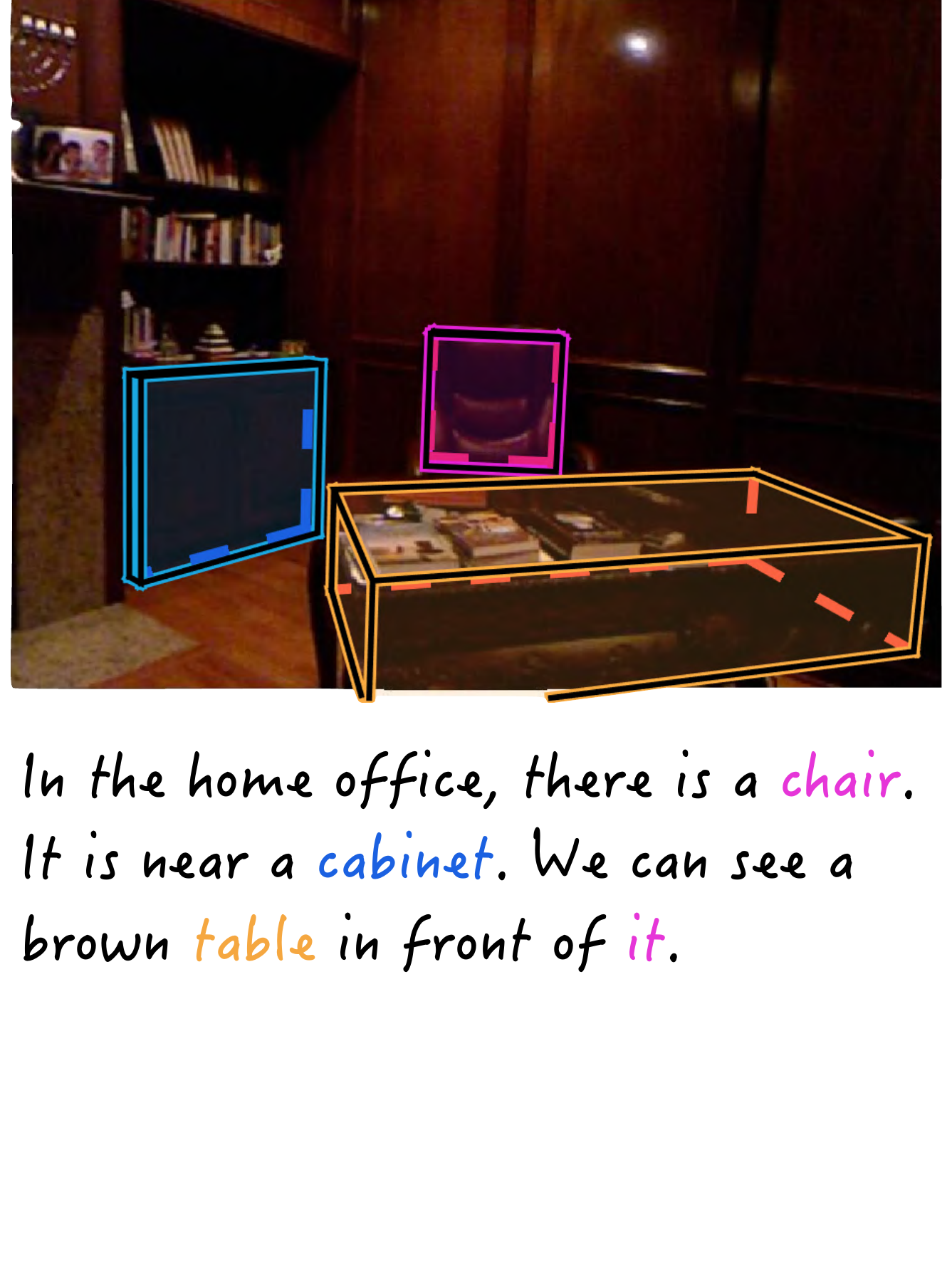}
    \includegraphics[height=\IHW,trim=105 270 105 60,clip=true]{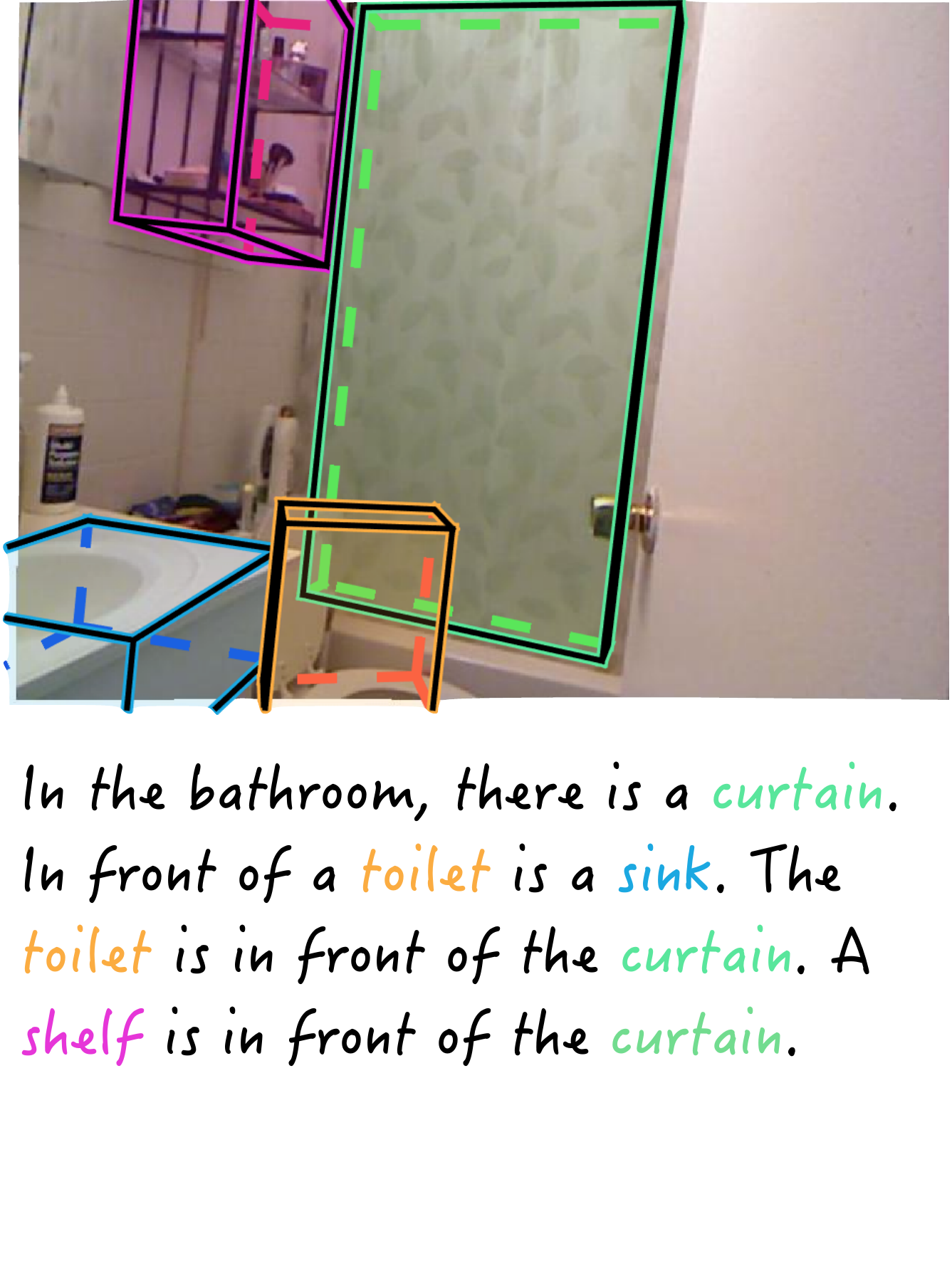}
       \includegraphics[height=\IHW,trim=105 270 105 60,clip=true]{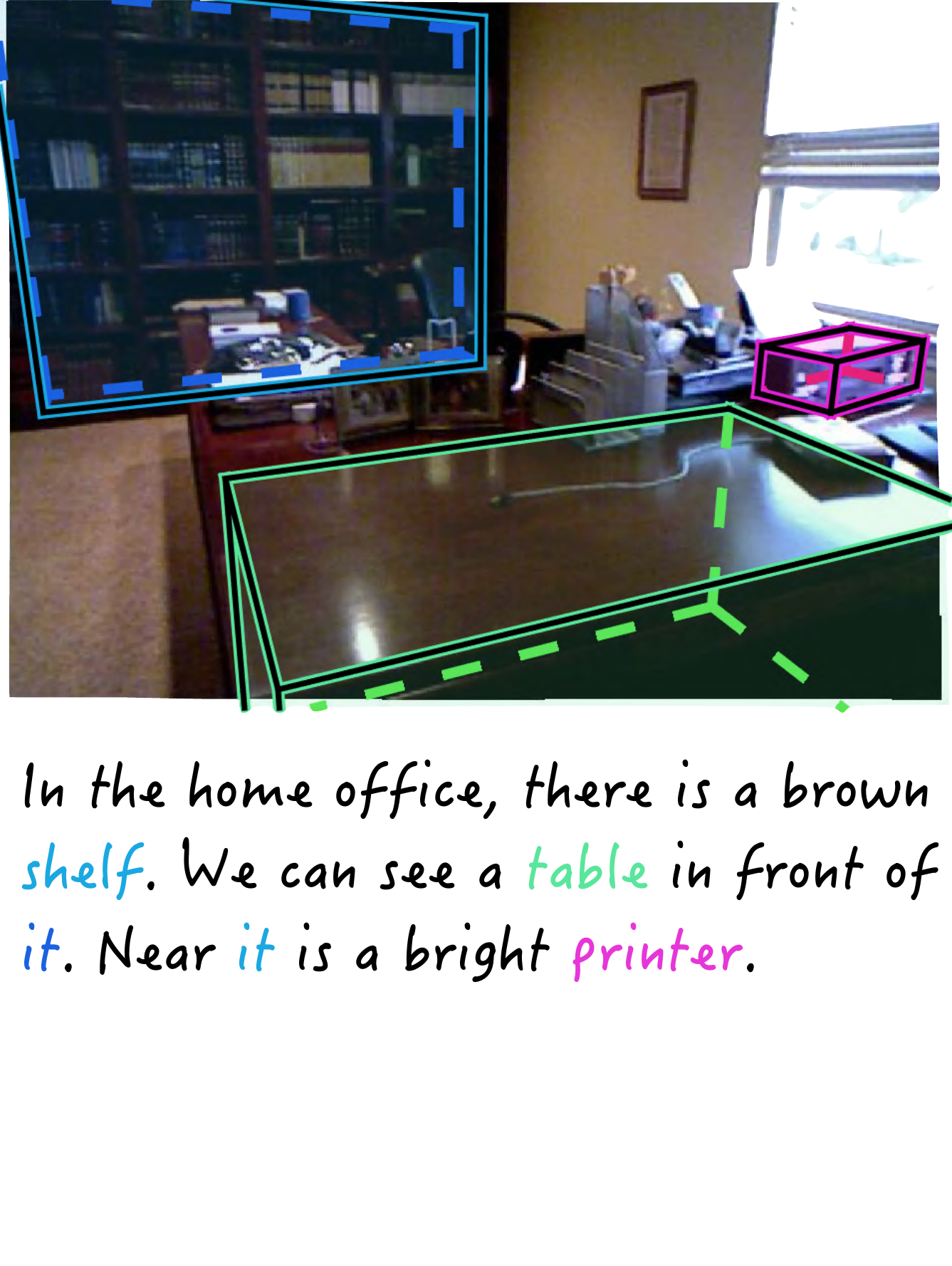}
    \\[-3.8mm]
    \includegraphics[height=\IHH,trim=105 270 105 60,clip=true]{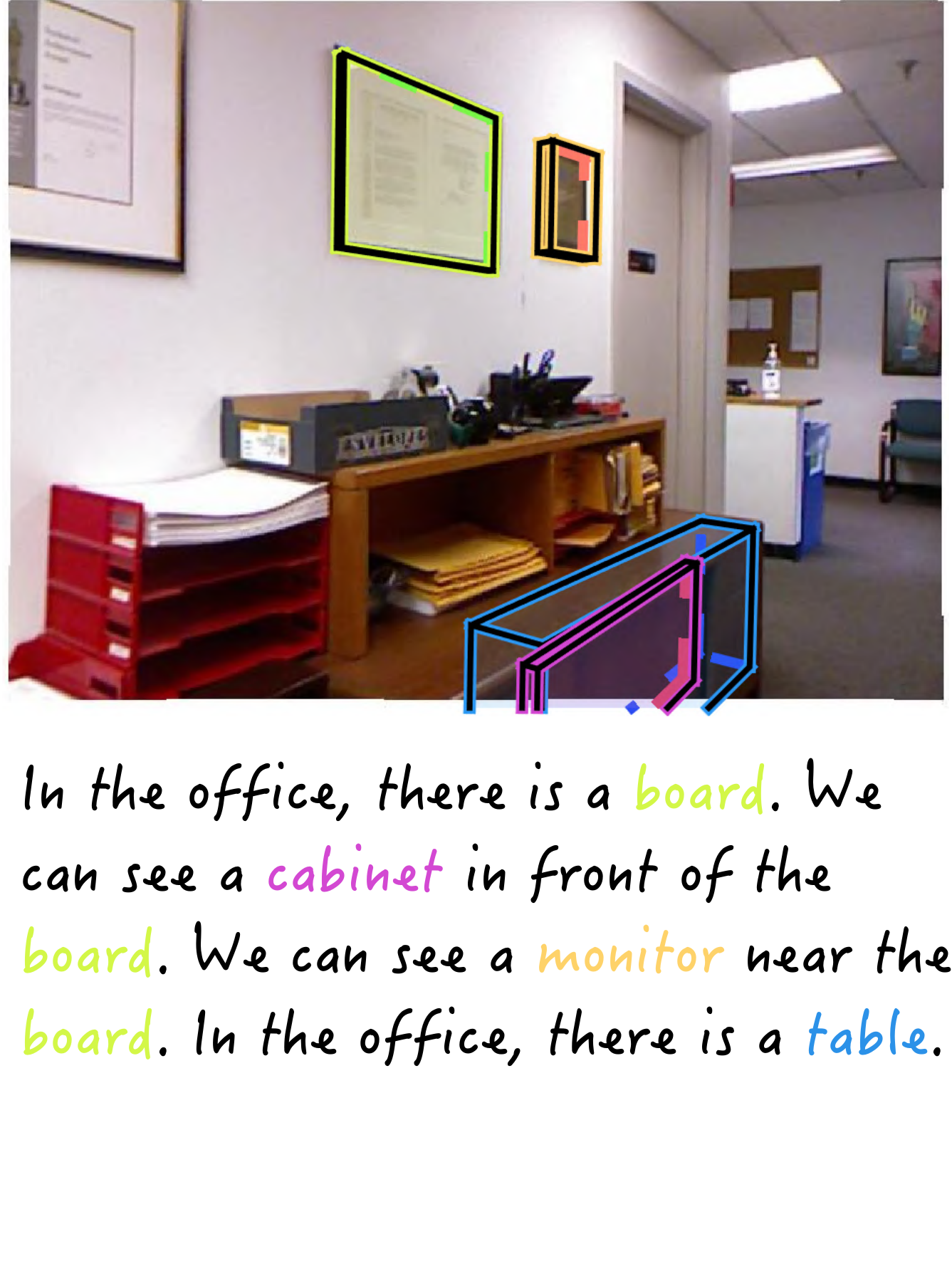}
    \includegraphics[height=\IHH,trim=105 270 105 60,clip=true]{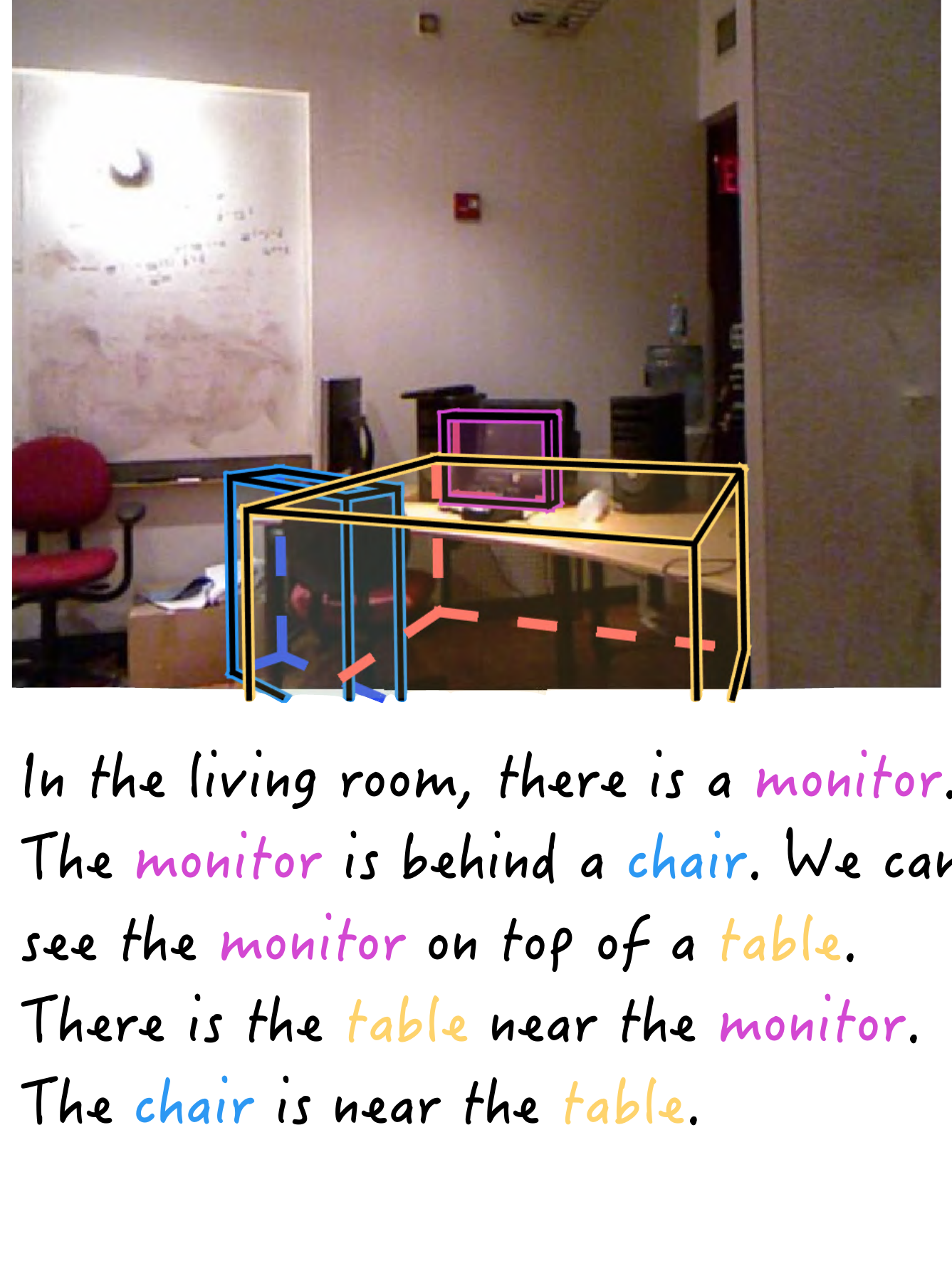}
    \includegraphics[height=\IHH,trim=105 270 105 60,clip=true]{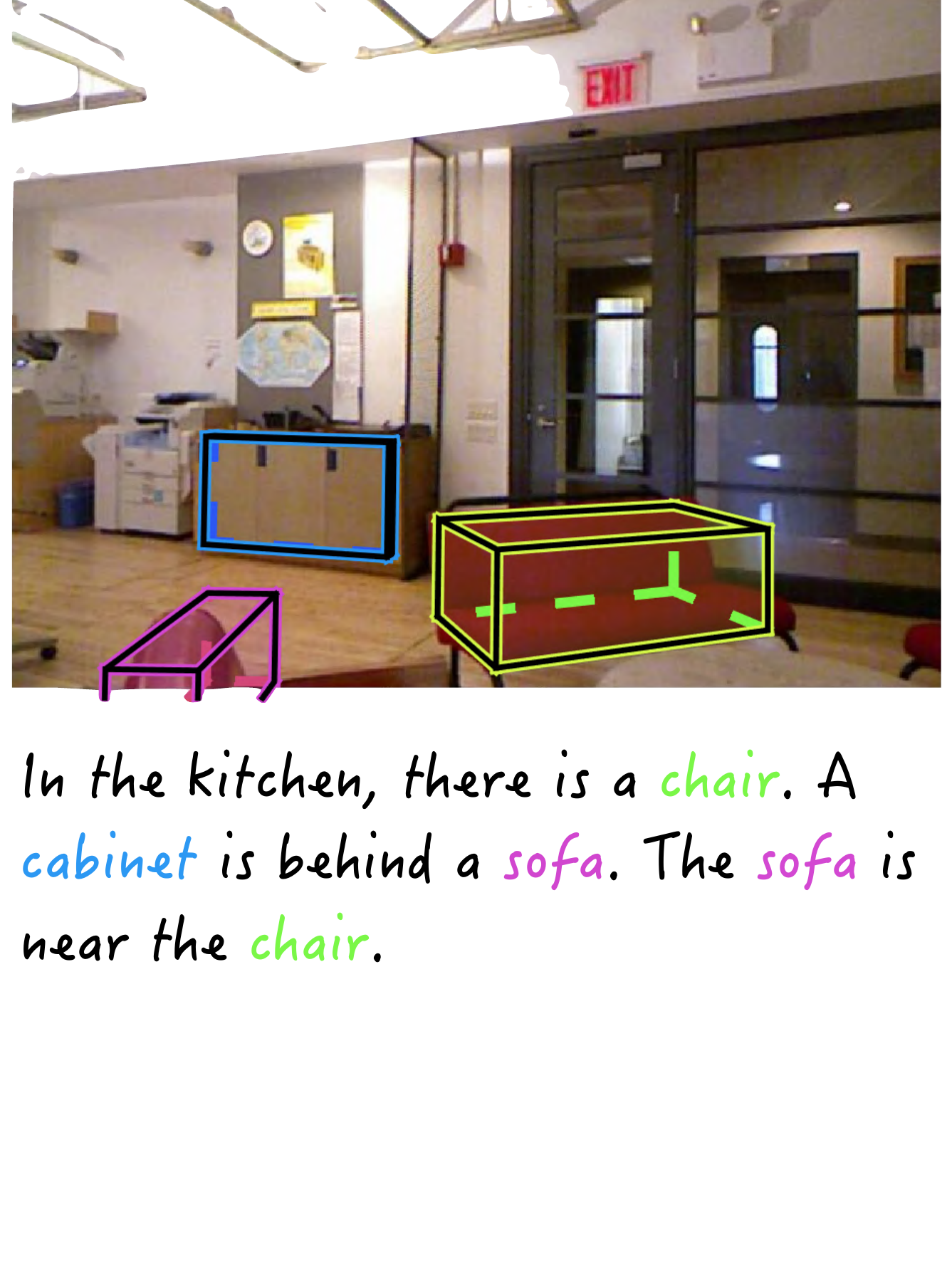}
     \includegraphics[height=\IHH,trim=105 270 100 60,clip=true]{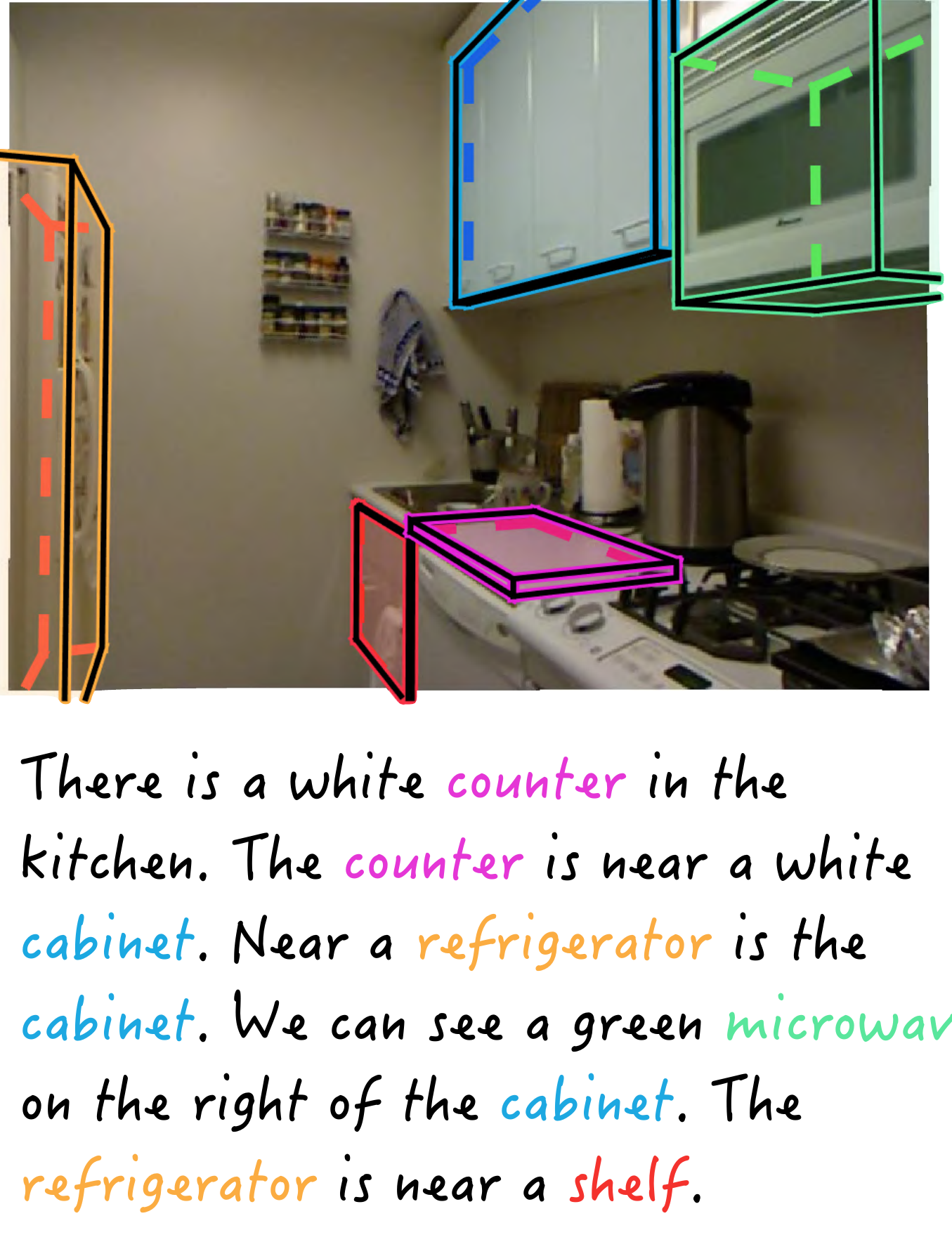}
     \vspace{-8mm}
    \caption{This Figure shows several examples of the descriptions generated using the proposed framework. In the top two rows the method builds on the ground-truth cuboids, while the bottom row shows the results using the  visual parser. Note that in the case of GT, the input to the method is the full set of GT objects for the image, thus the method still needs to take into account the saliency of what to talk about. We color-code object cuboids and nouns referring to them in text.}
    \label{fig:descrs}
\end{figure*}

\subsection{Performance Metrics}

To evaluate our method, we look for metrics typically used in machine translation. These include the BLEU~\cite{bleu} and ROUGE metrics among others. The BLEU score measures precision on $n$-grams, and is thus less suitable for our goal of lingual image description, as already noted in~\cite{mitchell12,Corso13}. On the other hand, ROUGE is an n-gram recall oriented measures which evaluates the information coverage between
summaries produced by the human annotators and those automatically produced by systems. ROUGE-1
(unigram) recall is the best option to use for comparing descriptions based only on predicted keywords~\cite{Corso13}.
ROUGE-2 (bigram) and ROUGE-SU4 (skip-4 bigram) are best to evaluate summaries
with respect to coherence and fluency. We use the ROUGE metrics following~\cite{Corso13} who uses it to evaluate lingual video summarization.

\subsection{Comparison of Results}

The proposed text generation method has five optional switches, controlling whether the following features are used during generation: 
(1) \texttt{diversity}: encourage diversity of the sentences by suppressing the entities and relations that have been mentioned; 
(2) \texttt{saliency}: draw salient objects with higher probability;
(3) \texttt{scene}: leading sentence mentions the class of the scene;
(4) \texttt{attributes}: use colors and sizes to describe objects when they are available;
(5) \texttt{coreference}: use a pronoun to refer to an object when it is mentioned in the previous sentence.
Our experiments test the framework under six feature-levels, level-0 to level-5, where the level-$k$ configuration uses the first $k$ features when generating the sentences. In particular, level-$0$ uses none of the features above, and thus each sentence is generated independently using the grammar; while level-$5$ uses all of these features.  

To put our performance in perspective, we compare our method to an intelligent baseline which follows a conventional approach in description generation. The baseline  describes an image by retrieving visually the most similar image from the training set, and simply using its corresponding description.
To compute our baseline, we use a battery of visual features such as spatial pyramids of SIFT, HOG, LBP, geometric context, etc, and kernels with different distances. We use~\cite{xiao10} to compute the kernels. Based on a combined kernel, we simply retrieve the training image with the highest matching score. 

Table~\ref{tab2} shows the results. We evaluate two settings: using ground-truth objects (denoted with GT) and using the results obtained via the visual parser (denoted with Real). We can see that the proposed method significantly outperforms the baseline in all three ROGUE measures. Also, configurations above level 3 are clearly better than level 1 and 2, which indicates that a special leading sentence that gives an overview of the scene is important for description generation. In addition, we observe that there are is a noticeable improvement from level 3 to level 4 and 5. This is not surprising: whereas attributes and coreference improve the quality of descriptions by making them richer and less verbose, such improvement on quality does not contribute substantially to the ROGUE score that are based on n-gram comparisons.  

Figure~\ref{fig:descrs} shows descriptions generated using our approach on a diverse set of scenes. It can be seen that linguistic issues such as sentence diversity, using attributes to describe objects, and using pronouns for coreferences have been properly addressed. However, there remain some problems that need future efforts to address. For example, since the choices of templates for different sentences are independent, sometimes an unfortunate selection of a template sequence may make the paragraph slightly unnatural.

\section{Conclusion}

We presented a new framework for generating natural descriptions of indoor scenes. Our framework integrates a CRF model for visual parsing, a generative grammar automatically learned from training descriptions, as well as a transformation algorithm to derive semantic trees from scene graphs, which takes into account the dependencies across sentences. Our experiments show substantially better descriptions than those produced by a baseline. Such findings indicate that high quality description generation requires not only reliable image understanding, but also delicate attention to linguistic issues, such as diversity, coherence, and logical order of sentences.

\clearpage

{
\bibliographystyle{acl}
\bibliography{biblio}
}

\end{document}